\documentclass[table]{article}

\usepackage{arxiv}

\usepackage[utf8]{inputenc} % allow utf-8 input
\usepackage[T1]{fontenc}    % use 8-bit T1 fonts
\usepackage{hyperref}       % hyperlinks
\usepackage{url}            % simple URL typesetting
\usepackage{booktabs}       % professional-quality tables
\usepackage{amsfonts}       % blackboard math symbols
\usepackage{amssymb}
\usepackage{amsmath}
\usepackage{nicefrac}       % compact symbols for 1/2, etc.
\usepackage{microtype}      % microtypography
\usepackage{lipsum}		% Can be removed after putting your text content
\usepackage{graphicx}
\usepackage[numbers]{natbib}
\usepackage{doi}
\usepackage{float}
\usepackage[capitalize]{cleveref}
\usepackage{subcaption}
\usepackage{xspace}
\usepackage{microtype}
\usepackage{tabularx}
\newcolumntype{Y}{>{\centering\arraybackslash}X}
\usepackage{highlight}
\usepackage{hhline}
\renewcommand{\opacity}{50}
\usepackage{multirow}
\usepackage{siunitx}

\newcommand*{\eg}{\emph{e.g.}\@\xspace}
\newcommand*{\ie}{\emph{i.e.}\@\xspace}
\newcommand\scs{\scriptsize{}}
\let\svthefootnote\thefootnote
\newcommand\freefootnote[1]{%
  \let\thefootnote\relax%
  \footnotetext{#1}%
  \let\thefootnote\svthefootnote%
}

\title{Multi-attribute balanced sampling for disentangled GAN controls}

\author{{Perla Doubinsky} \\
	\texttt{perla.doubinsky@lecnam.net} \\
	\And
	{Nicolas Audebert}\\
	\texttt{nicolas.audebert@cnam.fr} \\
\And
	{Michel Crucianu}\\
	\texttt{michel.crucianu@cnam.fr} \\
	\And
	{Hervé Le Borgne}\\
	\texttt{herve.le-borgne@cea.fr} \\
}
\date{}

\begin{document}
\maketitle

\begin{abstract}
Various controls over the generated data can be extracted from the latent space of a pre-trained GAN, as it implicitly encodes the semantics of the training data. The discovered controls allow to vary semantic attributes in the generated images but usually lead to entangled edits that affect multiple attributes at the same time. 
Supervised approaches typically sample and annotate a collection of latent codes, then train classifiers in the latent space to identify the controls. Since the data generated by GANs reflects the biases of the original dataset, so do the resulting semantic controls. 
We propose to address disentanglement by subsampling the generated data 
%to balance attribute co-occurrences and remove spurious correlations before training the classifiers. 
to remove over-represented co-occuring attributes thus balancing the semantics of the dataset before training the classifiers.
We demonstrate the effectiveness of this approach by extracting disentangled linear directions for face manipulation on two popular GAN architectures, PGGAN and StyleGAN, and two datasets, CelebAHQ and FFHQ. We show that this approach outperforms state-of-the-art classifier-based methods while avoiding the need for disentanglement-enforcing post-processing.
\end{abstract}

% keywords can be removed
\freefootnote{This paper is currently under consideration at Pattern Recognition Letters.}

\section{Introduction}
\label{sec:intro}

Generative Adversarial Networks (GANs)~\citep{gans2014goodfellow} produce high-resolution and photorealistic images by learning a mapping between a latent space, modelled by a random distribution, and the real image space. New images can then be obtained by randomly sampling in the latent space and feeding the latent codes to the generator. While it is easy to generate an image, its semantic properties might not be the desired ones. In applications such as data augmentation, it could be very useful to finely control the semantic properties of a generated image, especially to synthesize images that are difficult to capture in practice.

Recent research aim at leveraging pre-trained unconditional GANs and exploring their latent space to uncover the controls they can provide over the generated data. In particular, some methods find linear directions that can be interpreted as variations of some semantic attributes across the latent space \citep{harkonen2020ganspace, Plumerault2020Controlling, Jahanian2020steerability, interfacegan_bis, zhuang2021enjoy, yang2019semantic, Shen_2021_CVPR, pmlr-v119-voynov20a}.
However, the discovered directions often do not allow disentangled edits, affecting multiple attributes instead of solely altering the desired one.
Learning-based supervised methods commonly rely on a three-stages pipeline that consists in sampling a set of latent codes, then labelling the latent codes from the corresponding images using pre-trained image classifiers and finally, extracting the directions. As GANs learn to approximate the real data distribution that carries different kinds of biases, the sampling stage leads to generating biased datasets
that can, in turn, affect the semantic directions. 
The third stage is often performed by training a linear classifier to separate latent codes corresponding to images with a desired attribute (positive set) from latent codes corresponding to images without the desired attribute (negative set). The direction controlling the attribute is then taken as the vector orthogonal to the classifier's decision boundary~\citep{denton2019detectingbias, interfacegan_bis, yang2019semantic}. 
Existing correlations among attributes in the generated data may cause the positive and negative sets of a target attribute to be strongly imbalanced in respect to other attributes, thus biasing the direction towards those attributes. 

Inspired by this observation, we propose to enforce disentanglement by learning the semantic directions on datasets that are free from bias. Specifically, after sampling and labelling the latent codes, we subsample the dataset to balance the attributes joint distributions and remove correlations. 

We apply our method in the latent space of GANs trained for face synthesis to identify semantic directions corresponding to facial attributes. We conduct experiments on two types of GAN architectures: PGGAN~\citep{karras2018progressive} and StyleGAN~\citep{Karras_2019_CVPR}, respectively pre-trained on CelebAHQ~\citep{liu2015faceattributes} and FFHQ~\citep{Karras_2019_CVPR}. We provide a quantitative and qualitative comparison with the popular framework InterFaceGAN~\citep{interfacegan_bis}. We show that our approach leads to directions that are naturally disentangled whereas InterFaceGAN requires a post-processing step to reduce entanglement. 
We also show that, instead of relying on linear classifiers, directly using the direction connecting class centroids can give meaningful attribute controls for well-balanced data.

\section{Related work}

Early works on GANs uncovered some level of semantic structure in the latent space \eg by applying vector arithmetic on the latent codes \citep{DBLP:journals/corr/RadfordMC15}. 
Subsequent works focused on finding global directions in latent space corresponding to specific factors of variation ranging from geometric transformations (\eg position, scale) \citep{Jahanian2020steerability, Plumerault2020Controlling, spingarn2021gan}, memorability~\citep{Goetschalckx_2019_ICCV} to facial attributes~\citep{interfacegan_bis, Shen_2020_CVPR, harkonen2020ganspace, pmlr-v119-voynov20a, spingarn2021gan, zhuang2021enjoy, Shen_2021_CVPR}. By varying the latent codes towards those directions, the corresponding semantic properties of a generated image can be modified.
Recent proposals argue that semantics distribute non-linearly and locally~\citep{abdal_styleFlow, hou2020guidedstyle, Wang_2021_CVPR} but such methods are more expensive as they require to compute a specific manipulation for each input.

\textbf{Unsupervised methods.} Some works attempt to find semantic directions with self-supervised learning~\citep{pmlr-v119-voynov20a}, unsupervised approaches in latent space such as PCA~\citep{harkonen2020ganspace}, or by leveraging the internal representation of GANs to derive closed-form solutions~\citep{Shen_2021_CVPR, spingarn2021gan}.
However, since the semantics associated with each direction have to be manually identified afterwards, the discovery of the directions of interest is not guaranteed. In contrast, supervised methods aim to find directions corresponding to specific transformations \emph{a priori}.

\textbf{Supervised methods.} These methods typically sample a large number of latent codes, then annotate the corresponding synthesized images with semantic labels using pre-trained image classifiers \citep{interfacegan_bis, denton2019detectingbias, yang2019semantic, Wang_2021_CVPR, hou2020guidedstyle, abdal_styleFlow} to obtain a set of pairs (latent code, semantic labels). 
This set can be employed to train linear classifiers and each semantic direction is defined as the normal vector to the classifier decision boundary~\citep{denton2019detectingbias, interfacegan_bis, yang2019semantic}. 
The latent codes are sampled according to the latent space prior (usually a multivariate Gaussian), which transfers to the semantic directions the bias of the dataset used to train the generator. In contrast, we propose a subsampling method to obtain a collection of latent codes that is balanced w.r.t.\ multiple attributes and doesn't carry strong correlations, thus mitigating the propagation of bias. 

\textbf{Disentanglement of semantics.} Ideally, each of the discovered directions should control a single semantic property of the images. But very often the relation between directions and semantic properties is not one-to-one, \ie one direction has an impact on several properties; one speaks of \emph{entanglement}.
To reduce entanglement, some propose to refine the semantic directions afterwards, by enforcing an orthogonality constraint for the new directions. This post-processing step is referred to as ``conditional manipulation'' in~\citep{interfacegan_bis, Wang_2021_CVPR}.
Spingarn~\textit{et al}.~\citep{spingarn2021gan} introduce more constrained nonlinear paths that are defined as small circles on a sphere. Other works argue that entanglement is reduced if the transformations are learned together~\citep{zhuang2021enjoy, abdal_styleFlow}. 
For style-based GAN architectures, Hou~\textit{et al}.~\citep{hou2020guidedstyle} propose to learn an attention mechanism to manipulate the latent code for a particular layer.
Differently from previous work, our method addresses entanglement \emph{a priori} by debiasing the data employed to discover the directions. Hence, we argue that it can be complementary to previous proposals.

\begin{figure}[t]
\begin{subfigure}[t]{0.30\textwidth}
\includegraphics[width=1.1\textwidth]{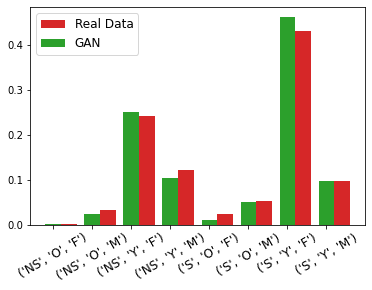}
\caption{
CelebAHQ vs.\ generated data with PGGAN CelebAHQ.}
\end{subfigure}
\quad
\begin{subfigure}[t]{0.30\textwidth}
\includegraphics[width=1.1\textwidth]{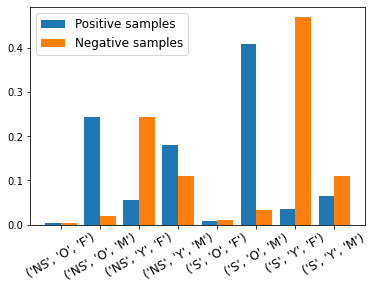}
\caption{
Positives vs.\ negatives w.r.t. ``Glasses'' for random sampling.}
\end{subfigure}
\quad
\begin{subfigure}[t]{0.30\textwidth}
\includegraphics[width=1.1\textwidth]{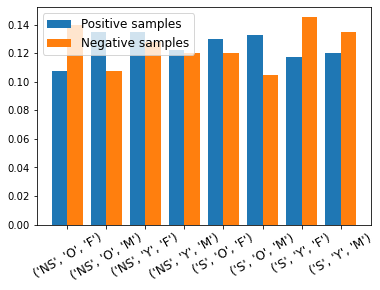}

\caption{
Positives vs.\ negatives w.r.t. ``Glasses'' for our sampling.}
\end{subfigure}
\caption{Joint distributions for three binary facial attributes ``Age'' (`O': Old, `Y': Young), ``Gender'' (`M': Male, `F': Female) and ``Smile'' (`S': Smile, `NS': No Smile). 
In (b), the positive set contains a majority of \emph{old males} while the negative set contains a majority of \emph{young females}, leading to bias the direction ``glasses'' toward the attributes ``age'' and ``gender''.}
\label{fig:data_bias}
\end{figure}

\section{Balanced sampling and direction estimation}\label{sec:our_method}

Let us consider a pre-trained generator $G(.)$ that maps a latent code $\mathbf{z}$ sampled from a $d$-dimensional latent space $\mathcal{Z} %\mathcal{L}
\subseteq \mathbb{R}^d$ to an image $\mathbf{I} = G(\mathbf{z})$ in image space $\mathcal{I} \subseteq \mathbb{R}^{H\times W\times C}$. Suppose the images are described by a set of binary attributes $\mathcal{A} = \{a_j, 1\leq j\leq m\}$. For each attribute $a_j$ we aim to find a global linear direction in the latent space, defined by unit vector $\mathbf{u}_j \in \mathbb{R}^d$, that allows to modify attribute $a_j$, and \emph{only} attribute $a_j$, in a generated image by translating the corresponding latent code $\mathbf{z}$ in that direction, $\mathbf{z'} = \mathbf{z} + \alpha \mathbf{u}_j$, $\alpha \in \mathbb{R}$ being the moving step. 

To find the directions, the procedure put forward in~\cite{Shen_2020_CVPR, yang2019semantic} is: (i)~train a multi-attribute image classifier $F_{\mathcal{I}}$ %(or $m$ single-attribute image classifiers $F_{\mathcal{I},j}$) 
on the ground truth provided with the database (\eg CelebA~\cite{liu2015faceattributes}); (ii)~generate $N$ latent codes and corresponding images $\{(\mathbf{z}_i,G(\mathbf{z}_i))_{i=1}^N\}$; (iii)~label every image %in $\mathcal{S_{\mathcal{I}}}$ 
with the %image 
classifier and associate the labels to the latent codes to produce $\mathcal{S} = \{(\mathbf{z}_i,F_{\mathcal{I}}(G(\mathbf{z}_i)))_{i=1}^N\}$; (iv)~for each attribute $j$, train a linear classifier $\Psi_j$ %$F_{\mathcal{L},j}$ 
in latent space on the $\mathcal{S}_j^+$ and $\mathcal{S}_j^-$ sets obtained from $\mathcal{S}$ by only considering the positive and respectively negative labels for attribute $j$. 
The direction in latent space allowing to control attribute $j$ is then defined by $\mathbf{u}_j$ the unit vector that is orthogonal to the decision boundary of the linear classifier $\Psi_j$. %$F_{\mathcal{L},j}$.

\subsection{Multi-attribute balanced sampling}
\label{sec:balancing}

The distribution of the binary attributes for a set of data can be represented in an $m$-dimensional contingency table (one dimension per attribute) where each of the $2^m$ cells contains the number of samples that have the corresponding combination of values for the $m$ attributes. If there are strong correlations between attributes in the GAN training data then the contingency table for that data is strongly imbalanced. The data in $\mathcal{S}$, generated by the trained GAN, is expected to show similar correlations. The example in~\cref{fig:data_bias}~(a) reveals that three attributes in the CelebA~\cite{liu2015faceattributes} dataset are strongly correlated (some combinations are much more frequent than others) and this reflects well in the random sample generated by the GAN\footnote{Other attributes in CelebA are also strongly correlated.}. For an attribute $a_j$, the sets $\mathcal{S}_j^+$ and $\mathcal{S}_j^-$ employed for training a classifier in the latent space mirror the imbalance in $\mathcal{S}$. If we consider the attribute ``Glasses'' in CelebA, \cref{fig:data_bias}~(b) shows how imbalanced the associated $\mathcal{S}_j^+$ and $\mathcal{S}_j^-$ sets are with respect to the three attributes in \cref{fig:data_bias}~(a). It is natural to expect that the classifier $\Psi_j$ %$F_{\mathcal{L},j}$ 
trained on such imbalanced data is influenced by the strong correlations. And, consequently, the unit vector $\mathbf{u}_j$ that is orthogonal to its decision boundary entangles the control of the target attribute with the most correlated attributes.

The idea of the method we propose is simple: subsample the data in $\mathcal{S}$ so as to obtain approximately the same number of samples in each cell of the contingency table. By removing the correlations from this table, we expect to strongly reduce the entanglement.

%Formulation avec 2 samplings
More precisely, we build a multi-attribute balanced sample $\mathcal{B}\subset \mathcal{S}$ by iteratively selecting data from $\mathcal{S}$ until we reach the total number of samples $N_0 \leq N$ we aim to obtain. At each iteration, we first uniformly sample one combination of attribute values (one cell of the contingency table), then we uniformly sample without replacement one data point $(\mathbf{z},F_{\mathcal{I}}(G(\mathbf{z})))$ with that combination. %of attribute values among the data not already considered.
In this way, at the end of the sampling procedure, we expect to have a balanced contingency table for $\mathcal{B}$ where each of the $2^m$ cells contains approximately $\frac{N_0}{2^m}$ data points, as shown in \cref{fig:data_bias}~(c). %Naturally, this also balances the marginals.
%Formulation avec 1 sampling
%More precisely, we build a multi-attribute balanced sample $\mathcal{B}$ of size $N_0$ by taking, for each combination of attribute values (or each cell of the contingency table), a uniform sample of size $\frac{N_0}{2^m}$ among the data points of $\mathcal{S}$ having that combination of attribute values. In this way, at the end of the sampling procedure, we expect to have a balanced contingency table for $\mathcal{B}$ where each of the $2^m$ cells contains $\frac{N_0}{2^m}$ data points. %Naturally, this also balances the marginals.

The subsampling procedure works well if there is enough data in $\mathcal{S}$ for each combination of attribute values. 
%Formulation avec 2 samplings
For strongly imbalanced data, we may have to address the case where there is no more data in $\mathcal{S}$ for one or more combinations before reaching the desired total number of samples $N_0$. 
%Formulation avec 1 sampling
%Since this is not the case for strongly imbalanced data, we have to address the situation where for some combinations $\mathcal{S}$ contains less than $\frac{N_0}{2^m}$ data points.
Note that, as we show in \cref{sec:experiments}, good results can be obtained with moderate values for $N_0$. 
The ideal solution for having a balanced $\mathcal{B}$ is to expand $\mathcal{S}$ by generating more images with $G$. But this can be very expensive since, as we found, the imbalance of $\mathcal{S}$ reflects the imbalance of the training dataset. Hence, we may require the generation of a very large number of images to obtain one more image with a rare combination of attribute values.

%Formulation avec 2 samplings
The solution we adopt consists in simply skipping the current iteration if no more data is available for that combination. 
%Formulation avec 1 sampling
%The solution we adopt consists in uniformly sampling data from $\mathcal{S}$ for combinations with more than $\frac{N_0}{2^m}$ data points until reaching the desired total number of samples $N_0$.
The resulting $\mathcal{B}$ is no longer so well-balanced but, as we show in~\cref{subsec:nb_samples}, there is a graceful decay in performance. An alternative is to oversample the data corresponding to the rarest combinations of attribute values, \ie random sample \emph{with} replacement for a combination if its cell in the contingency table of $\mathcal{S}$ has much less than $\frac{N_0}{2^m}$ data points. As shown in \cref{subsec:nb_samples}, this makes the decay in performance yet more graceful.

\subsection{Direction estimation}
\label{sec:classification_model}

The sampling procedure we described leads to a sample $\mathcal{B}$ of size $N_0$ that is balanced w.r.t.\ all attributes. 
For each attribute $j$, two sets $\mathcal{B}_j^+$ of size $N_j^+ \approx \frac{N_0}{2}$ and $\mathcal{B}_j^-$ of size $N_j^- \approx\frac{N_0}{2}$ can be readily obtained by considering the data having positive and respectively negative labels for attribute $j$. To find the direction $\mathbf{u}_j$ in latent space that allows to control attribute $j$, a good solution is to train a linear classifier on $\mathcal{B}_j^+\cup \mathcal{B}_j^-$, then take as $\mathbf{u}_j$ the vector orthogonal to the decision boundary. Preference is usually given (\eg~\cite{interfacegan_bis}) to linear Support Vector Machines (SVMs) that are fast to train and effective in high dimensions. To improve generalization, the value of the regularization hyperparameter could be selected by cross-validation. But as we find later in \cref{subsec:svm_vs_centroids}, when the dataset is balanced, a stronger regularization (larger SVM margin) tends to produce directions that allow more disentangled edits. If the linear SVM has a very large margin, the decision boundary becomes orthogonal to the line connecting the centroids of the two classes. For attribute $a_j$, this direction is defined by: %~\cref{eq:centroids_line}:
\begin{equation}\label{eq:centroids_line}
    \mathbf{u}_j = \frac{1}{N_j^{+}}\sum_{i=1}^{N_j^{+}} \mathbf{z}_i^{+} - \frac{1}{N_j^{-}}\sum_{i=1}^{N_j^{-}} \mathbf{z}_i^{-}, \quad \mathbf{z}^{+} \in B^{+}_j  \text{ and }  \mathbf{z}^{-} \in B^{-}_j.
\end{equation}
Experiments in \cref{subsec:svm_vs_centroids} show that entanglement is further reduced when this easy-to-compute direction is used to control the corresponding attribute. 

\section{Experiments}
\label{sec:experiments}

We evaluate and compare our proposal with the state-of-the-art method InterFaceGAN~\cite{interfacegan_bis}, considering the same attributes ``glasses'', ``gender'', ``smile'' and ``age''. For InterFaceGAN, the corresponding attribute control directions respectively produce the following effects: wearing glasses, presenting as male, smiling and getting younger. 
\cref{subsec:disentanglement} provides a detailed quantitative analysis of the effect a direction has for different attributes. The impact of sample size is evaluated in~\cref{subsec:nb_samples}, while in~\cref{subsec:svm_vs_centroids} we study regularization. Identity preservation is assessed in~\cref{subsec:id_preservation}.- Finally, qualitative results are shown in \cref{subsec:qualitative_results}.

\textbf{Models.} We conduct experiments with state-of-the-art GAN models trained on two face datasets, PGGAN CelebAHQ~\cite{karras2018progressive} and StyleGAN FFHQ~\cite{Karras_2019_CVPR}, generating $1024\times1024$ images (experiments with StyleGAN FFHQ trained on 256 $\times$ 256 images are shown in the supplementary material). Following~\cite{interfacegan_bis}, we train an auxiliary classifier
on CelebA~\cite{liu2015faceattributes} with a ResNet-50~\cite{He_2016_CVPR} using multi-task learning to predict the  attributes simultaneously. For each attribute, the task is a bi-classification problem with a softmax cross-entropy loss. We ensure that the accuracy of the classifiers is above 80\% (see details in the supplementary material).

\textbf{Implementations details.} We synthesize $N=1M$ images with PGGAN CelebAHQ and $N=500K$ images with StyleGAN FFHQ. We prepare a larger dataset for PGGAN as some combinations of attributes are rarer in CelebAHQ than in FFHQ. We apply the attribute predictors to all the generated images and discard the samples having a confidence below 0.9. For each attribute, we collect $N_0=1000$ samples using our multi-attribute balanced sampling. We choose this value depending on the number of samples in the cell with fewest samples (contingency tables are given in the supplementary material). 
The semantic directions are then obtained by taking the direction defined by the centroids of each class (see~\cref{sec:classification_model}).
For a fair comparison, we reproduce InterFaceGAN results instead of using the provided directions as they were not computed using the same attribute prediction model\footnote{The model was not made available by the authors.} nor the same number of samples.
For InterFaceGAN, we uniformly subsample the generated dataset then train linear SVMs with $C=1.0$ \footnote{As in the code provided by the authors: \url{https://github.com/genforce/interfacegan}} to obtain the semantic directions given by unit vectors. Since the dimension of the latent spaces of PGGAN and StyleGAN is $512$, these vectors are also $512d$.

\textbf{Metrics.} As in~\cite{interfacegan_bis}, we use the re-scoring metric to quantify the desired effect and entanglement associated with a direction. This metric measures how the attribute scores vary after manipulating the latent codes. Intuitively, a good direction should induce an increase in the score corresponding to the target attribute while not affecting other scores.
Given a direction $\mathbf{u}_j$ corresponding to attribute $a_j$, the re-scoring for attribute $a_k$ is computed as: 
\begin{equation}
    \Delta \mathbf{s}_{k}
=\frac{1}{n}\sum_{i=1}^n \left[F_{\mathcal{I},k}(G(\mathbf{z}_i))-F_{\mathcal{I},k}(G(\mathbf{z}_{i}+\alpha \mathbf{u}_j))\right]
\end{equation}
The desired \emph{effect} of direction $\mathbf{u}_j$ is given by the re-scoring result for the target attribute $a_j$ (higher is better). The \emph{entanglement} of direction $\mathbf{u}_j$ with another attribute $a_k$ is given by the re-scoring for that attribute (lower is better). We also derive a metric based on re-scoring to obtain the \emph{overall entanglement} associated with a direction. Similarly to StyleSpace~\cite{Wu_2021_CVPR}, we average the re-scoring results over the non-target attributes: $\frac{1}{|\mathcal{A}|-1}\sum_{i\in \mathcal{A}\backslash a_j}|\Delta \mathbf{s}_i|$. 
To quantify how the manipulations affect face identity, we employ a popular face recognition model pre-trained on VGGFace2~\cite{vgg_face2} and compute the cosine similarity between face embeddings before and after editing, as in~\cite{zhuang2021enjoy}. We extract embeddings of dimension 2048.

Both metrics are evaluated on $n=2000$ latent codes, we employ $\alpha=0.2$ for the editing and we report averages over 5 experiments.

\subsection{Disentanglement analysis}
\label{subsec:disentanglement}

PGGAN is a traditional GAN architecture where a code is sampled from a Gaussian latent space $\mathcal{Z}$ and fed to the first convolutional layer. In addition to $\mathcal{Z}$, StyleGAN introduces an intermediate latent space $\mathcal{W}$ whose distribution is modelled by fully-connected layers and learned during training, leading to a less entangled space~\cite{Karras_2019_CVPR}.
We compare our method to InterFaceGAN before (IfGAN) and after conditional manipulation (IfGAN + conditional), the latter having been introduced as an \emph{ad hoc} disentanglement post-processing~\cite{interfacegan_bis}. For attribute $j$, it consists in replacing $\mathbf{u}_j$ by the its projection on the subspace orthogonal to the directions found for the other attributes.

\textbf{PGGAN.} \cref{rescoring_pggan}~(a) provides the re-scoring results for InterFaceGAN without conditional manipulation. 
We observe that the diagonal scores increase after manipulating the latent codes, which shows that the directions have the desired effect. On the other hand, some of the off-diagonal scores also increase, indicating
entanglement with other attributes. For instance, the direction ``glasses'' also affects the attributes ``gender'' and ``age''. As shown in \cref{rescoring_pggan}~(b), the conditional manipulation allows to reduce the entanglement while maintaining the desired effect. According to \cref{rescoring_pggan}~(c), our approach succeeds to extract directions allowing disentangled edits without requiring conditional manipulation. It outperforms significantly InterFaceGAN and performs slightly better than conditional manipulation. 

\textbf{StyleGAN.} \cref{rescoring_stylegan_Z} shows the results in $\mathcal{Z}$ space. Compared to PGGAN there is less entanglement, probably because FFHQ is a larger dataset and the attributes are less correlated than in CelebAHQ. Otherwise, we find similar tendencies. In addition, the directions extracted with our approach have a stronger effect and are either on par or significantly more disentangled than for InterFaceGAN, even with conditional manipulation.
The results in $\mathcal{W}$ space are given in~\cref{rescoring_stylegan_W}. 
The $\mathcal{W}$ space being less entangled than $\mathcal{Z}$, the results of InterfaceGAN are good (conditional manipulation is not necessary) and those of our method are similar.
At a lower resolution (StyleGAN $256\times256$), $\mathcal{W}$ is nevertheless less disentangled and our method significantly improves disentanglement w.r.t.\ InterfaceGAN (see supplementary material).

\begin{table}[t]
\setlength\tabcolsep{1pt}
\begin{subtable}{0.38\textwidth}
\begin{tabularx}{\textwidth}{rYYYY|}
        & \rotatebox{30}{Glasses} & \rotatebox{30}{Gender} & \rotatebox{30}{Smile} & \rotatebox{30}{Age} \\
Glasses & \gradient{0.39} & \gradient{0.34} & \gradientneg{0.06} & \gradientneg{0.29} \\
Gender & \gradient{0.09} & \gradient{0.50} & \gradientneg{0.06} & \gradientneg{0.22} \\
Smile & \gradientneg{0.03} & \gradientneg{0.07} & \gradient{0.37} & \gradientneg{0.02} \\
Age & \gradientneg{0.04} & \gradientneg{0.31} & \gradientneg{0.07} & \gradient{0.15}
\end{tabularx}
\caption{IfGAN~\cite{interfacegan_bis}
}
\end{subtable}
\begin{subtable}{0.30\textwidth}
\begin{tabularx}{\textwidth}{YYYY|}
\rotatebox{30}{Glasses}  & \rotatebox{30}{Gender} & \rotatebox{30}{Smile} & \rotatebox{30}{Age} \\
\gradient{0.27} & \gradient{0.10} & \gradientneg{0.02} & \gradientneg{0.10} \\
\gradientneg{0.00} & \gradient{0.41} & \gradientneg{0.03} & \gradientneg{0.02}\\
\gradientneg{0.02} & \gradientneg{0.04} & \gradient{0.36} & \gradientneg{0.01}\\
\gradientneg{0.02} & \gradientneg{0.11} & \gradientneg{0.04} & \gradient{0.13}\\
\end{tabularx}
\caption{IfGAN + conditional~\cite{interfacegan_bis}
}
\end{subtable}
\begin{subtable}{0.30\textwidth}
\begin{tabularx}{\textwidth}{YYYY}
\rotatebox{30}{Glasses} & \rotatebox{30}{Gender} & \rotatebox{30}{Smile} & \rotatebox{30}{Age} \\
 \gradient{0.32} & \gradient{0.07} & \gradientneg{0.05} & \gradientneg{0.07} \\
 \gradient{0.01} & \gradient{0.42} & \gradientneg{0.04} & \gradientneg{0.05} \\
 \gradientneg{0.01} & \gradientneg{0.02} & \gradient{0.36} & \gradientneg{0.01}\\
 \gradientneg{0.02} & \gradientneg{0.13} & \gradientneg{0.06} & \gradient{0.14}
\end{tabularx}
\caption{Our method}
\end{subtable}
\caption{Re-scoring results in $\mathcal{Z}$ space for PGGAN CelebAHQ. Each row shows the effect of a direction on all attributes, each column shows the effect of all directions on an attribute. Effect (diagonal values) should be high, entanglement (off-diagonal values) should be low.
}
\label{rescoring_pggan}
\end{table}

\begin{table}[t]
\setlength\tabcolsep{1pt}
\begin{subtable}{0.38\textwidth}
\begin{tabularx}{\textwidth}{rYYYY|}
        & \rotatebox{30}{Glasses} & \rotatebox{30}{Gender} & \rotatebox{30}{Smile} & \rotatebox{30}{Age} \\
Glasses & \gradient{0.36} & \gradient{0.23} & \gradientneg{0.05} & \gradientneg{0.19} \\
Gender & \gradient{0.16} & \gradient{0.37} & \gradientneg{0.11} & \gradientneg{0.18} \\
Smile & \gradient{0.03} & \gradientneg{0.08} & \gradient{0.15} & \gradient{0.00} \\
Age & \gradientneg{0.12} & \gradientneg{0.25} & \gradient{0.00} & \gradient{0.18} \\
\end{tabularx}
\caption{IfGAN~\cite{interfacegan_bis}}
\end{subtable}
\begin{subtable}{0.30\textwidth}
\begin{tabularx}{\textwidth}{YYYY|}
\rotatebox{30}{Glasses}  & \rotatebox{30}{Gender} & \rotatebox{30}{Smile} & \rotatebox{30}{Age} \\
\gradient{0.27} & \gradient{0.12} & \gradientneg{0.01} & \gradientneg{0.08} \\
\gradient{0.06} & \gradient{0.29} & \gradientneg{0.07} & \gradientneg{0.07}\\
\gradientneg{0.04} & \gradientneg{0.06} & \gradient{0.15} & \gradient{0.00} \\
\gradientneg{0.07} & \gradientneg{0.15} & \gradient{0.00} & \gradient{0.15} \\
\end{tabularx}
\caption{IfGAN + conditional~\cite{interfacegan_bis}}
\end{subtable}
\begin{subtable}{0.30\textwidth}
\begin{tabularx}{\textwidth}{YYYY}
\rotatebox{30}{Glasses} & \rotatebox{30}{Gender} & \rotatebox{30}{Smile} & \rotatebox{30}{Age} \\
\gradient{0.35} & \gradient{0.06} & \gradientneg{0.02} & \gradientneg{0.05} \\
\gradient{0.02} & \gradient{0.33} & \gradientneg{0.08} & \gradientneg{0.06}\\
\gradientneg{0.01} & \gradientneg{0.05} & \gradient{0.17} & \gradient{0.00} \\
\gradientneg{0.07} & \gradientneg{0.13} & \gradientneg{0.01} & \gradient{0.17} \\
\end{tabularx}
\caption{Our method}
\end{subtable}
\vspace{8pt}
\caption{Re-scoring results in $\mathcal{Z}$ space for StyleGAN FFHQ.}
\label{rescoring_stylegan_Z}
\end{table}

\begin{table}[t]
\setlength\tabcolsep{4pt}
\begin{subtable}{0.48\textwidth}
\begin{tabularx}{\textwidth}{rYYYY}
        & \rotatebox{0}{Glasses} & \rotatebox{0}{Gender} & \rotatebox{0}{Smile} & \rotatebox{0}{Age} \\
Glasses & \gradient{0.51} & \gradient{0.09}  & \gradient{0.02} & \gradientneg{0.09} \\
Gender & \gradient{0.08} & \gradient{0.39} & \gradientneg{0.22} & \gradientneg{0.06} \\
Smile & \gradient{0.09} & \gradientneg{0.14} & \gradient{0.22} & \gradientneg{0.03} \\
Age & \gradientneg{0.11} & \gradientneg{0.11} & \gradientneg{0.04} & \gradient{0.20} \\
\end{tabularx}
\caption{IfGAN~\cite{interfacegan_bis}}
\end{subtable}
\begin{subtable}{0.48\textwidth}
\begin{tabularx}{\textwidth}{rYYYY}
& \rotatebox{0}{Glasses} & \rotatebox{0}{Gender} & \rotatebox{0}{Smile} & \rotatebox{0}{Age} \\
Glasses & \gradient{0.63} & \gradient{0.09} & \gradientneg{0.07} & \gradientneg{0.04} \\
Gender & \gradient{0.05} & \gradient{0.44} & \gradientneg{0.10} & \gradientneg{0.08} \\
Smile & \gradient{0.05} & \gradientneg{0.06} & \gradient{0.22} & \gradientneg{0.05} \\
Age & \gradientneg{0.09} & \gradientneg{0.15} & \gradientneg{0.01} & \gradient{0.20} \\
\end{tabularx}
\caption{Our method}
\end{subtable}
\caption{Re-scoring results in $\mathcal{W}$ space for StyleGAN FFHQ.}
\label{rescoring_stylegan_W}
\end{table}

\subsection{Impact of the sample size}
\label{subsec:nb_samples}

\begin{figure}[t]
\includegraphics[width=\textwidth]{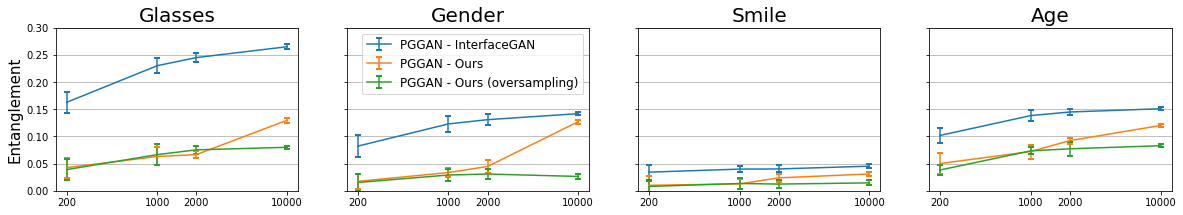}
\includegraphics[width=\textwidth]{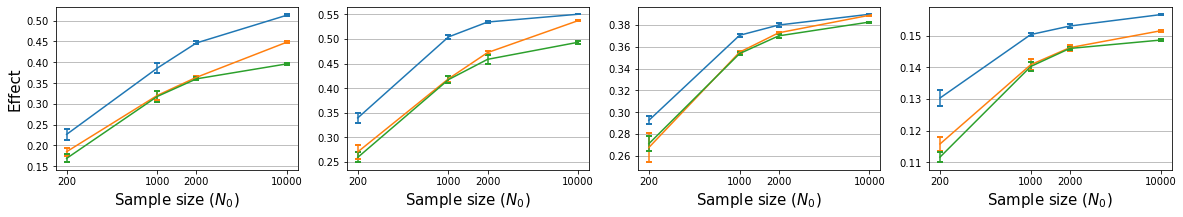}
\caption{Influence of sample size on the overall entanglement (top) and desired effect (bottom) associated with a direction in the $\mathcal{Z}$ space of PGGAN CelebAHQ.}
\label{fig:nb_samples_pgganZ}
\end{figure}

We study the impact of the sample size on the effect of the extracted directions. For our method, we consider both settings, \ie sampling without replacement and sampling with replacement (oversampling). 
Figure~\ref{fig:nb_samples_pgganZ} shows the results in $\mathcal{Z}$ space for PGGAN CelebAHQ (similar results are given for StyleGAN FFHQ in the supplementary material, for $\mathcal{Z}$ and $\mathcal{W}$). 
For both our method and InterFaceGAN, we find that the effect and the entanglement increase with the size of the sample. For moderate values of $N_0$ the distributions are well-balanced, hence the level of entanglement of our directions remains significantly below that of InterFaceGAN directions. However, we observe a significant increase for $N_0=\num{10000}$, suggesting that the distributions are no longer balanced as many cells of the contingency table have been emptied (see %contingency 
tables in the supplementary material). 
We find that oversampling allows to mitigate this effect. Additionally, the size of the sample has a limited impact regarding direction variability between runs. Standard deviations increase when we decrease the size of the sample but remain reasonably small. 
Following these observations, we argue that a large sample size (as in~\cite{interfacegan_bis}) is not necessary to obtain meaningful directions. Nevertheless, oversampling allows to increase sample size for a stronger effect, while keeping a low entanglement.

\subsection{SVM vs.\ centroids difference}
\label{subsec:svm_vs_centroids}

\begin{figure}[!t]
\includegraphics[width=\textwidth]{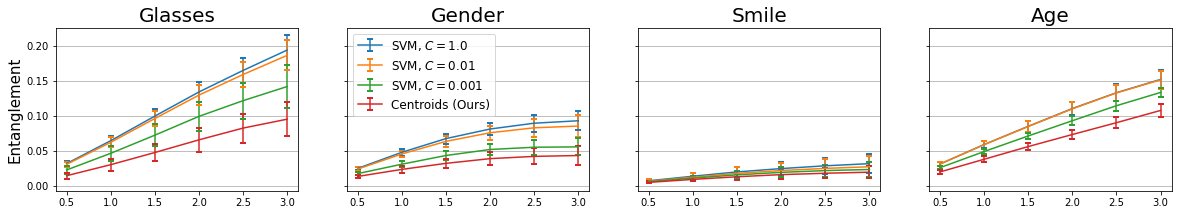}
\includegraphics[width=\textwidth]{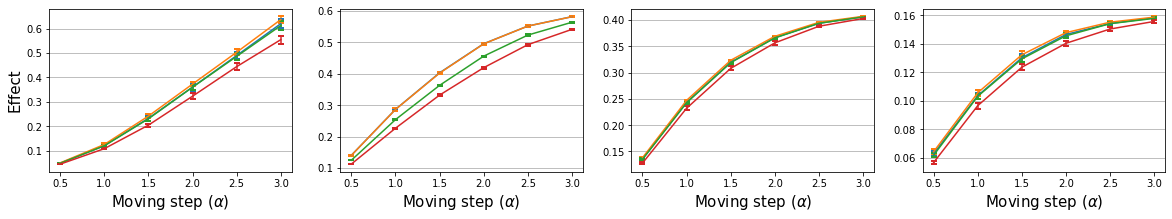}
\caption{Influence of SVM regularization on the overall entanglement (top) and desired effect (bottom) associated with a semantic direction in the $\mathcal{Z}$ space of PGGAN CelebAHQ.
}
\label{influence_regularization_pggan}
\end{figure}

For a balanced dataset obtained with the method described in \cref{sec:balancing}, \cref{influence_regularization_pggan} shows for the $\mathcal{Z}$ space of PGGAN CelebAHQ that a stronger regularization (smaller value of $C$) leads to smaller entanglement, while the effect on the target attribute remains almost unchanged. Similar results are reported in the supplementary material for the $\mathcal{Z}$ and $\mathcal{W}$ spaces of StyleGAN FFHQ. This observation led us to consider the case of a very large SVM margin, when the decision boundary becomes orthogonal to the direction connecting the centroids of the two classes (see~\cref{sec:classification_model}). We find that this direction gives the best performances.

\subsection{Identity preservation}
\label{subsec:id_preservation}

\cref{identity_preservation} shows a quantitative evaluation of identity preservation. We compare our results with InterFaceGAN for the different models. Some attributes affect the identity more than others, in particular ``gender''. For InterFaceGAN, we observe that the identity is less well-preserved when manipulating ``age'' and ``glasses'', which might be explained by the fact that they are entangled with ``gender'' (cf.\ Tables~\ref{rescoring_pggan}, \ref{rescoring_stylegan_Z}). For ``smile'', which is a well-disentangled attribute, we perform slightly better or on par with InterFaceGAN for directions with similar effects (cf.~Tables~\ref{rescoring_pggan}, \ref{rescoring_stylegan_Z}, \ref{rescoring_stylegan_W}). This suggests that our directions preserve the identity well. Our results in the $\mathcal{W}$ space of StyleGAN for the attributes ``glasses'' and ``age'' are quite below those of InterFaceGAN, which is probably due to our directions having more effect (see \eg~\cref{fig:qual_results_continuous_stylegan} where we end up with quite occluding sunglasses). 

\begin{table}[H]
%    \centering
    \begin{tabularx}{\textwidth}{r Y Y Y | Y Y Y | Y Y Y | Y Y Y}
        \toprule
        & & Glasses & & & Gender & & & Smile & & & Age & \\
        \midrule
         & P & S $\mathcal{Z}$ & S $\mathcal{W}$ & P & S $\mathcal{Z}$ & S $\mathcal{W}$ & P & S $\mathcal{Z}$ & S $\mathcal{W}$& P & S $\mathcal{Z}$ & S $\mathcal{W}$ \\
         \midrule
        IfGAN & 0.59 \scs{$\pm$ 0.14} & 0.72 \scs{$\pm$ 0.13}& \textbf{0.70} \scs{$\pm$ 0.11} & 0.56 \scs$\pm$ 0.14& 0.66 \scs$\pm$ 0.14& \textbf{0.71} \scs$\pm$ 0.13 & 0.74 \scs$\pm$ 0.10& 0.86 \scs$\pm$ 0.09 &  \textbf{0.81} \scs$\pm$ 0.09 & 0.66 \scs$\pm$ 0.15& 0.70 \scs$\pm$ 0.14& \textbf{0.76} \scs$\pm$ 0.11 \\
        Ours & \textbf{0.77} \scs{$\pm$ 0.15}& \textbf{0.73} \scs{$\pm$ 0.13} & 0.63 \scs{$\pm$ 0.12} & \textbf{0.62} \scs$\pm$ 0.14 & \textbf{0.74} \scs$\pm$ 0.13 & 0.68 \scs$\pm$ 0.13 & \textbf{0.79} \scs$\pm$ 0.08 & \textbf{0.87} \scs$\pm$ 0.08 & 0.78 \scs$\pm$ 0.08 & \textbf{0.78} \scs$\pm$ 0.10& \textbf{0.73} \scs$\pm$ 0.12& 0.69 \scs$\pm$ 0.11 \\
         \bottomrule
    \end{tabularx}
    \caption{Identity preservation (higher is better) for manipulation in $\mathcal{Z}$ of PGGAN CelebAHQ (P) and in $\mathcal{Z}$ and $\mathcal{W}$ of StyleGAN FFHQ (S $\mathcal{Z}$ resp. S $\mathcal{W}$).}
    \label{identity_preservation}
\end{table}

\subsection{Qualitative results}
\label{subsec:qualitative_results}

\begin{figure}
    \centering
    \begin{tabular}{cccc|cccc}
    Glasses & Gender & Smile & Age & Glasses & Gender & Smile & Age \\
    \includegraphics[width=0.11\textwidth]{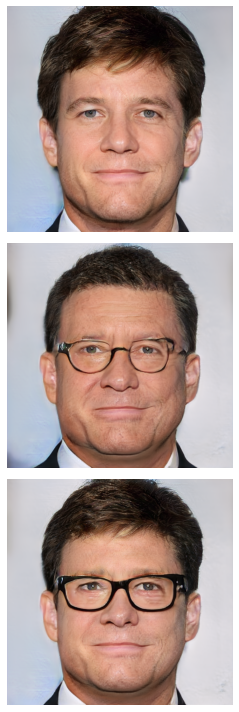} & 
    \includegraphics[width=0.11\textwidth]{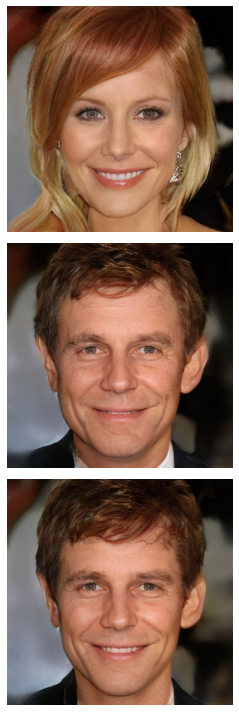} &
    \includegraphics[width=0.11\textwidth]{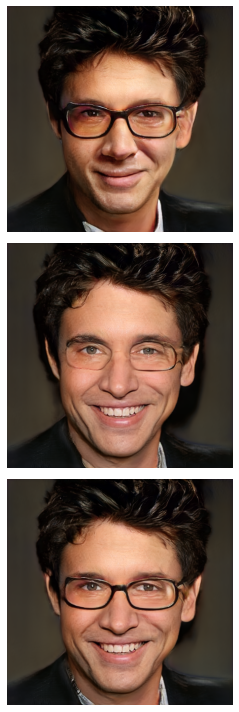} &
    \includegraphics[width=0.11\textwidth]{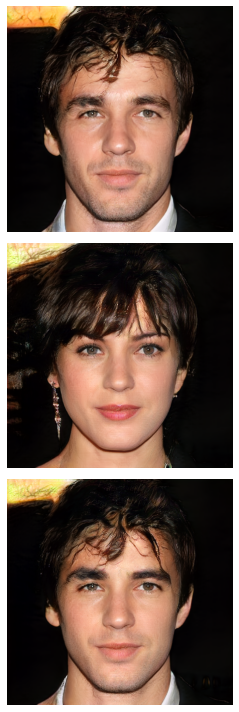} & 
    \includegraphics[width=0.11\textwidth]{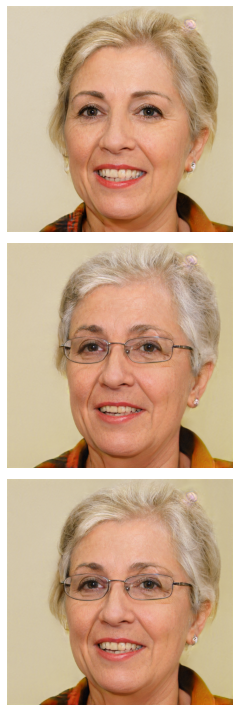} &
    \includegraphics[width=0.11\textwidth]{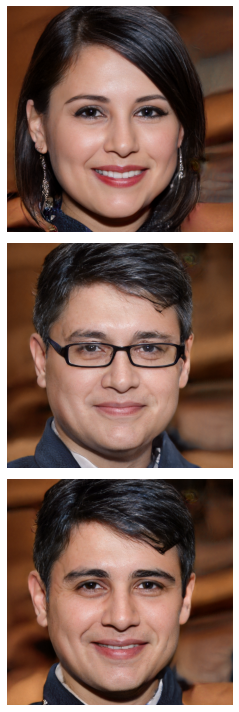} &
    \includegraphics[width=0.11\textwidth]{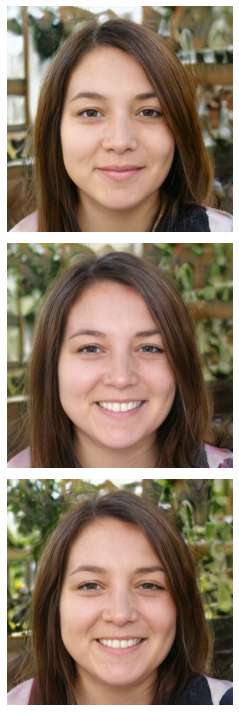}&
    \includegraphics[width=0.11\textwidth]{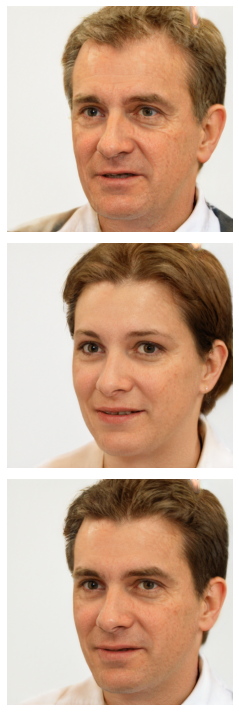}\\
    \end{tabular}
    \caption{Qualitative results in $\mathcal{Z}$ space of PGGAN CelebAHQ (left) and StyleGAN FFHQ (right). First row: input image. Second row: InterFaceGAN. Third row: our method.}
    \label{fig:qualitative_results}
\end{figure}

\begin{figure}[t]
    \centering
    \rotatebox[origin=c]{90}{\small IfGAN} \raisebox{-0.4\height}{\includegraphics[width=0.9\textwidth]{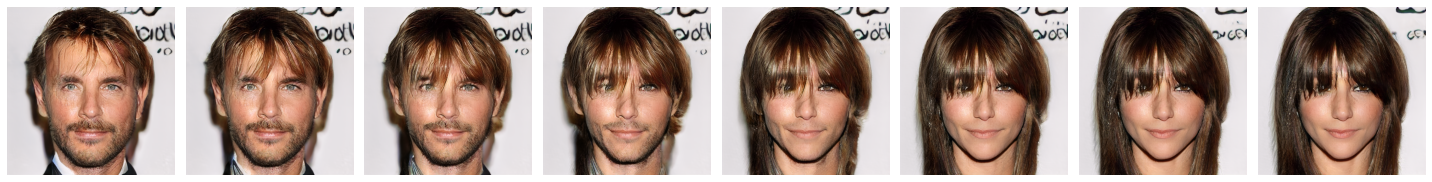}} \\
    \rotatebox[origin=c]{90}{\small Ours} \raisebox{-0.4\height}{\includegraphics[width=0.9\textwidth]{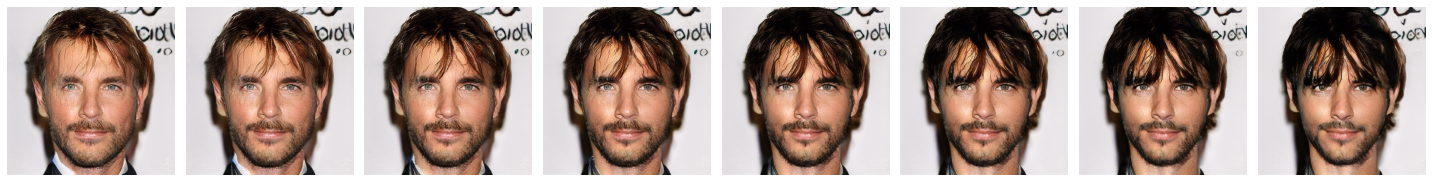}} 
    \caption{Continuous manipulation for the attribute ``age'' in $\mathcal{Z}$ space of PGGAN CelebAHQ.}
    \label{fig:qual_results_continuous_pggan}
\end{figure}

\begin{figure}[h!]
    \centering
    \rotatebox[origin=c]{90}{\small IfGAN} \raisebox{-0.4\height}{\includegraphics[width=0.9\textwidth]{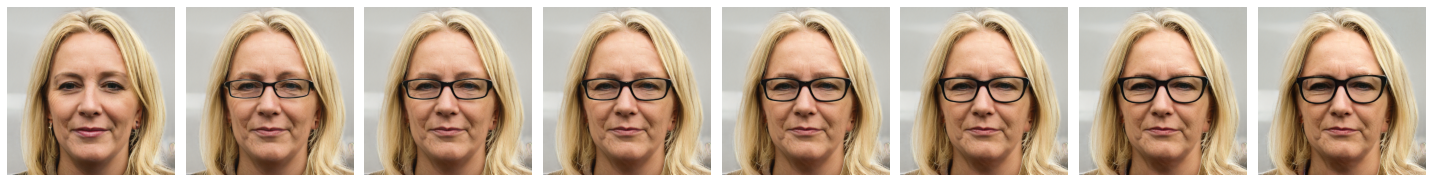}} \\
    \rotatebox[origin=c]{90}{\small Ours} \raisebox{-0.4\height}{\includegraphics[width=0.9\textwidth]{age_celebahq_pggan_continuous_ours.png}}
    \caption{Continuous manipulation for the attribute ``glasses'' in $\mathcal{W}$ space of StyleGAN FFHQ.}
    \label{fig:qual_results_continuous_stylegan}
\end{figure}

In \cref{fig:qualitative_results}, we show qualitative results in the $\mathcal{Z}$ space of PGGAN CelebAHQ and StyleGAN FFHQ. We find that the images obtained with InterFaceGAN show significant entanglement. On par with the quantitative results, we notice that the direction ``glasses'' is entangled with ``age'' and ``gender'', the direction ``gender'' also affects the attributes ``age'' and ``glasses'' and the direction ``age'' tends to feminize. 
In contrast, our directions better preserve the non-target attributes. 
In~\cref{fig:qual_results_continuous_pggan} and \cref{fig:qual_results_continuous_stylegan} we show qualitative results for different amplitude values. Additional results for the different attributes are presented in the supplementary material.

\section{Discussion}\label{sec:conclusion_discussion}

While our method balances the sample to decorrelate the attributes, we observed that the resulting directions in latent space are quasi-orthogonal (see supplementary material), which was not \textit{a priori} expected. 
This may explain the success of previous works that look for orthogonal directions in the latent space. For example, GANSpace~\cite{harkonen2020ganspace} applies PCA in the $\mathcal{W}$ space and the authors are able to assign semantic interpretations to the resulting directions (orthogonal by definition). The conditional manipulation in InterFaceGAN~\cite{interfacegan_bis} also enforces an orthogonality constraint among control directions to reduce entanglement. This requirement of orthogonality did not have an \emph{a priori} justification but our results indicate that orthogonality in latent space could be a necessary condition for independent controls and, even for unconditional GANs, the latent space does encode a significant part of the semantics. 
We believe that our subsampling approach can prove beneficial to other works on GAN control that rely on sampling in the latent space. Two issues could be raised. First, as in most works on finding supervised controls, we use pseudo-labels provided by image classifiers that are assumed reliable. But they can also be affected by bias, with an impact on both the labelling of the training set and the evaluation since re-scoring depends on the classifiers. However, results on FFHQ show that even classifiers trained on smaller datasets like CelebAHQ transfer quite well. Second, using classifiers to find directions assumes that samples can be grouped in classes. This nevertheless works surprisingly well for continuous attributes that are binarized (\eg ``age'') and might not be a problem in practice.

\section{Conclusion}

We focused on the identification of linear directions in the latent space of a GAN to control semantic attributes of the generated images. Our assumption was that the entanglement typically observed in such situations results from strong correlations among attributes in the training data, that are transferred to the generated data. To address this issue, we proposed a simple and general method that balances the data among the different combinations of values for the attributes. The evaluation on two popular GAN architectures and two face datasets shows that this approach outperforms state-of-the-art classifier-based methods while avoiding the need for post-processing.

%\nocite{ren2021do}
%\nocite{khrulkov2021on}

\bibliographystyle{unsrt}
\bibliography{biblio.bib, biblio_loc.bib}

\end{document}